# Conditions Under Which Conditional Independence and Scoring Methods Lead to Identical Selection of Bayesian Network Models

Robert G. Cowell
Department of Actuarial Science and Statistics
City University, London
Northampton Square
London EC1V 0HB, UK.

## Abstract

It is often stated in papers tackling the task of selecting a Bayesian network structure from data that there are these two distinct approaches: (i) Apply conditional independence tests when testing for the presence or otherwise of edges; (ii) Search the model space using a scoring metric.

Here I argue that for complete data and a given node ordering this division is largely a myth, by showing that cross entropy methods for checking conditional independence are mathematically identical to methods based upon discriminating between models by their overall goodness-of-fit logarithmic scores.

## 1  Introduction.

In this paper I consider learning Bayesian network structures on a finite set of discrete variables, under the restrictions of complete data and a given node-ordering. The following quote (Cheng et al. 1997) is typical of statements made in articles either introducing a novel algorithm or reviewing current algorithms for learning Bayesian networks.

> Generally, these algorithms can be grouped into two categories: one category of algorithms uses heuristic searching methods to construct a model and then evaluates it using a scoring method. ... The other category of algorithms constructs Bayesian networks by analyzing dependency relationships between nodes.

While on the face of it these two approaches appear quite different, I will argue that model search methods based upon maximizing a local log-score can be expressed as equivalent search methods employing local conditional independence tests.

The plan of the paper is as follows. The next section introduces notation together with some theoretical results. Section 3 states the assumptions made in later sections. Section 4 considers learning structure from a known distribution, which is equivalent to learning from an infinite data set. Section 5 considers the more realistic case of inferring model structure from finite data, from both a classical and a Bayesian perspective.

## 2  Notation and background results.

I will assume that readers are familiar with the notion of a Bayesian network; for a recent monograph see Cowell et al. (1999). I consider a finite set $X = \{X_1, \ldots, X_n\}$ of $n$ (finite) discrete random variables taking values in the state space $\mathcal{X} = \{\mathcal{X}_1, \ldots, \mathcal{X}_n\} = \times_{i=1}^n \mathcal{X}_i$. If $A \subseteq V := \{1, \ldots, n\}$ denotes some index set, then $X_A$ will denote the subset of variables $\{X_a : a \in A\}$ and will take values in $\mathcal{X}_A := \times_{a \in A} \mathcal{X}_a$. Where convenient a variable $X_v$ may be referred to by its index $v$. Particular configurations will be denoted using lower case letters, for example, $x = (x_1, \ldots, x_n)$, or $x_A \in \mathcal{X}_A$.

In this paper I consider search algorithms constrained by a given node ordering; without loss of generality I will take the node ordering to be $(X_1, \ldots, X_n)$. Let $\mathcal{G}_n$ denote the set of directed acyclic graphs (dags) on $X$, such that $X_i$ can be a parent of $X_j$ only if $i < j$.

For $g \in \mathcal{G}_n$ let $\mathcal{P}_g$ denote the set of distributions directed Markov with respect to $g$. This means that for any $P_g \in \mathcal{P}_g$, the probability mass function factorizes

as

$$p_g(x) = \prod_{v \in V} p_g(x_v \mid x_{\text{pa}(v:g)}),  \quad (1)$$

where $X_{\text{pa}(v:g)}$ denotes the set of parents of the vertex $v$ in $g$.

Let $P(X)$ and $Q(X)$ be two probability distributions over $X$. The Kullback–Leibler divergence between $P$ and $Q$ is defined to be

$$K(P,Q) = \mathrm{E}_P\left[\log \frac{p(X)}{q(X)}\right] = \sum_{x \in \mathcal{X}} p(x) \log \frac{p(x)}{q(x)}.$$

It takes a non-negative value measuring the similarity or closeness of the distribution $Q$ to that of $P$, vanishing if and only if the distributions are identical.

It was shown by Cowell (1996) (see also Cowell et al. (1999)) that for a given graph $g \in \mathcal{G}_n$ the distribution $P_g \in \mathcal{P}_g$ which minimizes $K(P, P_g)$ for some fixed distribution $P(X)$ assigns to every vertex $v \in V$,

$$P_g(X_v \mid X_{\text{pa}(v:g)}) = P(X_v \mid X_{\text{pa}(v:g)}). \quad (2)$$

Let $A$, $B$ and $C$ be disjoint index subsets of $V$, and let $P(X)$ be some distribution over $X$. Then the *cross-entropy* of $X_A$ and $X_B$ is defined to be

$$H_P(X_A, X_B) = \mathrm{E}_P\left[\log \frac{p(X_A, X_B)}{p(X_A)p(X_B)}\right],$$

whilst the cross-entropy of $X_A$ and $X_B$ conditional on $X_C$, or *conditional cross entropy*, is defined as

$$H_P(X_A, X_B \mid X_C) = \\ \mathrm{E}_P\left[\log \frac{p(X_A, X_B \mid X_C)}{p(X_A \mid X_C)p(X_B \mid X_C)}\right].$$

We say $X_A$ is conditionally independent of $X_B$ given $X_C$ under the distribution $P$, written as $X_A \perp\!\!\!\perp_P X_B \mid X_C$, if and only if $p(X_A, X_B \mid X_C) = p(X_A \mid X_C)p(X_B \mid X_C)$ (Dawid 1979). The notation $\perp\!\!\!\perp_P$ will be abbreviated to $\perp\!\!\!\perp$ when the distribution $P$ under consideration is clear from the context. Note that if $X_A \perp\!\!\!\perp_P X_B \mid X_C$, then $H_P(X_A, X_B \mid X_C) = 0$, and vice versa.

For $g \in \mathcal{G}_n$ any distribution $P_g \in \mathcal{P}_g$ has the *directed Markov property*, that is, any node is conditionally independent of its non-descendants given its parents in $g$:

$$X_v \perp\!\!\!\perp X_{\text{nd}(v:g)} \mid X_{\text{pa}(v:g)}.$$

Note in particular that this implies $H_{P_g}(X_v, X_{\text{nd}(v:g)} \mid X_{\text{pa}(v:g)}) = 0$.

## 3  Assumptions made in learning a network.

I shall make the following assumptions for the remaining sections.

1. I will be looking for good predictive models, selected according to a log-scoring rule, and choosing the simplest model among equally good predictive models.

2. The dataset is complete, and there are no latent variables.

3. The node ordering is given, and without loss of generality is $(X_1, \ldots, X_n)$.

4. There are no logical constraints between the various conditional probability tables to be estimated.

Assumption 1 emphasizes that I am not looking to construct causal models from data, but simply seeking good predictive models. The log scoring rule is unique in that it is (for multi-attribute variables) the only proper scoring rule whose value is determined solely by the probability attached to the outcome that actually occurs (Bernardo 1979). The final part of Assumption 1 is Occam's razor: without it I could choose the saturated model, that is, the complete graph, which would fit the data perfectly. Scoring based search methods usually try and balance these two aspects — by penalizing a model's predictive score with some measure of the model's complexity — as a way of reducing overfitting.

Assumption 2 is made for simplicity, to avoid approximations being made to handle missing data, or having to account for the pattern of missing data. It also implies that the logarithmic score of a dag decomposes additively into functions (one function for each node, depending upon the node and its parents in the dag), thus making local search possible by enabling independent optimizations of each node's parent set.

Assumption 3 implies that the dag I obtain might not exhibit all of the conditional independence properties of the data, but only those consistent with the ordering.

Assumption 4 states that I am assuming local meta-independence of the conditional probabilities associated with the families of any given graph considered (Dawid and Lauritzen 1993). These conditional probabilities will be taken as parameters to be estimated.

## 4 Learning networks from a known distribution.

In this section, I assume that the joint distribution $P(X)$ is known; this is equivalent to recovering $P(X)$ from its maximum likelihood estimate (MLE) for the saturated model in the limit of an infinite amount of data drawn from $P(X)$. The task is to find the simplest model $g \in \mathcal{G}_n$ such that $P_g(X) = P(X)$.

### 4.1 Model selection via conditional independence tests.

Given the ordering $(X_1, \ldots, X_n)$, the joint distribution $P(X)$ may be factorized as

$$P(X) = \prod_{i=1}^{n} P(X_i \mid X_1, \ldots, X_{i-1}).$$

The goal of the model search using conditional independence tests is to find for each node $X_i$ a minimal set $X_{\text{pa}(i)} \subseteq \{X_1, \ldots, X_{i-1}\}$, such that

$$P(X_i \mid X_{\text{pa}(i)}) = P(X_i \mid X_1, \ldots, X_{i-1}), \quad (3)$$

which is equivalent to the independence statement

$$X_i \perp\!\!\!\perp \{X_1, \ldots, X_{i-1}\} \setminus X_{\text{pa}(i)} \mid X_{\text{pa}(i)}.$$

The minimal set $X_{\text{pa}(i)}$ may then be taken as the set of parents of node $X_i$ in the sought for graph. If found for each $X_i$, the joint distribution will factorize as (1). Let us write $R_i := \{X_1, \ldots, X_{i-1}\} \setminus X_{\text{pa}(i)}$, and $X_I = \{X_1, \ldots, X_i\}$. Then using the identity $P(X_i, R_i \mid X_{\text{pa}(i)}) = P(X_i \mid R_i, X_{\text{pa}(i)}) P(R_i \mid X_{\text{pa}(i)})$, we have

$$H_P(X_i, R_i \mid X_{\text{pa}(i)})$$
$$= \sum_{x_I} p(x_i, r_i, x_{\text{pa}(i)}) \log \left[ \frac{p(x_i, r_i \mid x_{\text{pa}(i)})}{p(x_i \mid x_{\text{pa}(i)}) p(r_i \mid x_{\text{pa}(i)})} \right]$$
$$= \sum_{x_I} p(x_i, r_i, x_{\text{pa}(i)}) \log \left[ \frac{p(x_i \mid r_i, x_{\text{pa}(i)})}{p(x_i \mid x_{\text{pa}(i)})} \right]. \quad (4)$$

When (3) holds the conditional cross entropy (4) vanishes, and conversely. Hence (4) forms the basis of a conditional independence test. Note that if $X_i \perp\!\!\!\perp R_i \mid X_{\text{pa}(i)}$, then for any subset $S_i \subset R_i$ it is also true that $X_i \perp\!\!\!\perp S_i \mid X_{\text{pa}(i)}$, and

$$H_P(X_i, S_i \mid X_{\text{pa}(i)}) =$$
$$\sum p(x_i, s_i, x_{\text{pa}(i)}) \log \left[ \frac{p(x_i \mid s_i, x_{\text{pa}(i)})}{p(x_i \mid x_{\text{pa}(i)})} \right] \quad (5)$$

will vanish also (and conversely). In principle, one could perform an exhaustive search over all possible sets $R_i$ to find the largest such set for which the cross entropy vanishes. In practice, this is not usually possible because the search space is too large. Thus heuristic searches are normally applied, usually based upon evaluating (5) with $S_i$ singleton sets. An *example* of such a search is:

1. Set $X_{\text{pa}(i)} := \emptyset$ and $R_i = \{X_1, \ldots, X_{i-1}\}$.

2. WHILE $X_i \perp\!\!\!\perp R_i \mid X_{\text{pa}(i)}$ is FALSE do
   - Select $S_i \in R_i$ such that $H_P(X_i, S_i \mid X_{\text{pa}(i)})$ is maximized.
   - Remove $S_i$ from $R_i$ and add it to $X_{\text{pa}(i)}$.

3. WHILE $\exists Y \in X_{\text{pa}(i)}$ such that $X_i \perp\!\!\!\perp Y \mid X_{\text{pa}(i)} \setminus \{Y\}$ is TRUE do
   - $X_{\text{pa}(i)} \to X_{\text{pa}(i)} \setminus \{Y\}$.

This is similar to the 'thickening and thinning' algorithm of Cheng et al. (1997). More generally, $S_i$ could represent a restricted set of subsets of $R_i$, not just singleton sets.

### 4.2 Model selection via Kullback–Leibler divergence.

Given a graph $g \in \mathcal{G}_n$, the distribution directed Markov with respect to $g$ (that is, factorizes as (1)) which has minimum Kullback–Leibler divergence from $P(X)$ obeys (2) for each node in the graph. In principle, one could perform an exhaustive search over all possible graphs $g \in \mathcal{G}_n$, finding their closest matching distributions $P_g(X)$ in terms of Kullback–Leibler divergence from $P(X)$, selecting those graphs for which the Kullback–Leibler divergence vanishes, and selecting among these graphs the one having the fewest number of edges.

Consider a graph $g \in \mathcal{G}_n$, and it associated distribution $P_g(X)$ which satisfies (2). The Kullback–Leibler divergence is given by

$$\sum_x p(x) \log \left[ \prod_i \frac{p(x_i \mid x_{I-1})}{p(x_i \mid x_{\text{pa}(i:g)})} \right] = \sum_i \sum_{x_I} p(x_I) \log \left[ \frac{p(x_i \mid x_{I-1})}{p(x_i \mid x_{\text{pa}(i:g)})} \right]. \quad (6)$$

Note that (6) decomposes into a sum of terms, one for each node, where the $g$-dependence of the term on each node depends upon the family in $g$ of that node. In fact for the same graph $g$, the $i$th term in the summation (6) is *identical* to the cross entropy expression (4). Thus, *an exhaustive search based upon conditional independence is equivalent to an exhaustive search which minimizes Kullback–Leibler divergence.*

Suppose in a stepwise search algorithm that $g$ is our current model and a candidate model $g'$ differs from $g$ in one node $X_i$ for which $X_{\text{pa}(i:g')} \supset X_{\text{pa}(i:g)}$. Then the difference in Kullback–Leibler divergence of the two models is found from (6) to be

$$\Delta(g,g') = \\ H_P(X_i, X_{\text{pa}(i:g')} \setminus X_{\text{pa}(i:g)} \mid X_{\text{pa}(i:g)}), \qquad (7)$$

which is (5) with $S_i := X_{\text{pa}(i:g')} \setminus X_{\text{pa}(i:g)}$. Thus choosing the $g'$ differing from $g$ by one or more edges which maximizes (7) is equivalent to choosing $g' \supset g$ which minimizes $K(P, P_{g'})$. After adding parents to $X_i$ until no further decrease in Kullback–Leibler divergence is possible (on adding yet further nodes as parents), one could thin the parents of node $X_i$, by removing nodes for which $\Delta(g,g')$ remains zero.

More generally, *a decision criterion in the search algorithm which moves to a model $g'$ from a submodel $g$ based upon the consideration of a set of possible sets $S_i$ and their associated conditional independence tests could be used to give the same result (or move) based upon the consideration of the same $S_i$ and the changes in Kullback–Leibler divergence, because the numerical quantities entering into the decision process upon which the decision is based are identical in the two search frameworks.*

Put another way, for every search heuristic based upon using conditional independence tests, there is an equivalent search heuristic based upon using changes in Kullback–Leibler divergence, and vice versa. There is no fundamental difference between the two approaches, only a difference in interpretation.

## 5  Learning networks from finite data

### 5.1  Model selection via conditional independence tests.

The directed Markov property, and the completeness of the data, allows conditional independence tests to be performed locally on each node. The conditional independence tests employ MLEs. However due to sampling variability the tests are not sharp, so typically the requirement of the exact vanishing of (conditional) cross entropy expressions is relaxed.

Thus for example, suppose that $g$ is our current model, with node $X_i$ having parents $X_{\text{pa}(i:g)}$, and from the data one obtains the MLEs $\hat{P}(X_i, X_{\text{pa}(i:g)})$. Furthermore, suppose that for some node or set of nodes $S_i \in X_{I-1} \setminus X_{\text{pa}(i:g)}$ one evaluates the MLEs $\hat{P}(X_i, X_{\text{pa}(i:g)}, S_i)$ and the conditional cross entropy

$$\sum_{x_i, x_{\text{pa}(i:g)}, s_i} \hat{p}(x_i, x_{\text{pa}(i:g)}, s_i) \times \\ \log\left[\frac{\hat{p}(x_i \mid x_{\text{pa}(i:g)}, s_i)}{\hat{p}(x_i \mid x_{\text{pa}(i:g)})}\right], \qquad (8)$$

where $\hat{p}(x_i \mid x_{\text{pa}(i:g)}) = \hat{p}(x_i, x_{\text{pa}(i:g)})/\hat{p}(x_{\text{pa}(i:g)})$ etc. Then a search heuristic would employ a decision rule which on the basis of the value of (8) would either accept or reject the hypothesis that $X_i \perp\!\!\!\perp S_i \mid X_{\text{pa}(i)}$, and if the latter, decide which among the candidate $S_i$ to add to $X_{\text{pa}(i)}$ for the next iteration of the search algorithm. Two common decision heuristics are: (i) if the value of (8) is below a fixed threshold value $\epsilon$ then accept the conditional independence; (ii) perform a classical significance test, using the null hypothesis of conditional independence, under which a suitable multiple (8) (the multiplier being twice the number of observations) will have a $\chi_k^2$ distribution for some suitable $k$. Each of these has a counterpart in search heuristics based upon a log-score.

### 5.2  Model selection via maximum likelihood.

Let us write $n(x_A)$ for the marginal count of the number of cases in the dataset for which $X_A = x_A$. For $g \in \mathcal{G}_n$ the log-likelihood of the data decomposes as

$$\log L(p_g) = \\ \sum_i \prod_{x_i, x_{\text{pa}(i:g)}} n(x_i, x_{\text{pa}(i:g)}) \log\left(p_g(x_i \mid x_{\text{pa}(i:g)})\right),$$

which yields the MLE

$$\hat{p}_g(x_i \mid x_{\text{pa}(i:g)}) = \frac{n(x_i, x_{\text{pa}(i:g)})}{n(x_{\text{pa}(i:g)})}. \qquad (9)$$

Suppose, as in Section 4.2, that $g$ is our current model and $g'$ differs from $g$ in one node $X_i$ for which $X_{\text{pa}(i:g')} \supset X_{\text{pa}(i:g)}$. Then the difference in the log-likelihoods of the two models evaluated at their MLEs is given by

$$\log \frac{L(\hat{p}_{g'})}{L(\hat{p}_g)} = \sum_{x_i, x_{\text{pa}(i:g')}} n(x_i, x_{\text{pa}(i:g')}) \times \\ \log \frac{n(x_i, x_{\text{pa}(i:g')})/n(x_{\text{pa}(i:g')})}{n(x_i, x_{\text{pa}(i:g)})/n(x_{\text{pa}(i:g)})}. \qquad (10)$$

Thus one could decide to move from $g$ to $g'$ in the model search if this quantity is positive. However, this will generally be the case with finite data, because the larger model will fit the data better by virtue of having extra parameters, hence the significance of the better fit needs assessing. One simple heuristic is to set a threshold $\epsilon$ such that if the change is greater than $\epsilon$ the difference is taken to be significant — to do this we must first normalize (10) by the total number of cases $N = \sum_x n(x)$ in the dataset. Doing this yields

$$\frac{1}{N} \log \frac{L(\hat{p}_{g'})}{L(\hat{p}_g)} = \sum_{x_i, x_{\mathrm{pa}(i:g')}} \hat{p}(x_i, x_{\mathrm{pa}(i:g')}) \log \frac{\hat{p}(x_i \mid x_{\mathrm{pa}(i:g')})}{\hat{p}(x_i \mid x_{\mathrm{pa}(i:g)})},$$

where $\hat{p}$ represents the (marginal of the) MLE of the saturated model. This is identical to (8), with $S_i = X_{\mathrm{pa}(i:g')} \setminus X_{\mathrm{pa}(i:g)}$.

A more formal approach would be based on hypothesis testing. Note that twice the value of (10) is the difference in the deviances of the two models, which under the assumption that the larger model is true, and that the smaller model is also true, will have a $\chi^2_k$ distribution with $k$ equal to the difference in the degrees of freedom of the two nested models. Thus we perform the same test, and obtain the same result, as the formal conditional independence test described at the end of Section 5.1. Alternatively, one could penalize the deviance by some function of the number of parameters, for example by using the Akaike Information Criterion (Akaike 1973) which penalizes the more complex model by twice the number of extra parameters.

More generally, because of the equality of (8) and (9) it follows as in the last paragraph of Section 4.2, that for every search heuristic based upon testing for conditional independence, there is an equivalent search heuristic based upon using changes in log-maximum-likelihood, and vice versa. There is no fundamental difference between the two approaches, only a difference in interpretation.

### 5.3 The Bayesian approach.

Many belief network search algorithms using a scoring metric tend to employ the Bayesian formalism, with the score being the log-marginal likelihood. The advantages are that for smaller data sets, where the asymptotic distribution results required for the tests in Section 5.1 and Section 5.2 may not apply (although exact classical tests are available, see Chapter 4 of Lauritzen (1996)), the results tend to be more robust and, furthermore, generally less sensitive to the presence of zeroes in marginal counts.

The Bayesian approach requires a prior on the space of graphical structures – usually this is taken to be uniform, but there are other alternatives (Heckerman 1998). For each graphical structure a prior on the probability parameters is also required – usually these are taken to be locally independent Dirichlet priors.

Under these assumptions and complete data the marginal likelihood may be evaluated explicitly and decomposes into a product of terms, one for each node. An early and important paper is Cooper and Herskovits (1992), who gave an explicit formula for the marginal likelihood under these conditions.

A common feature of the analyses given in Section 5.1 and Section 5.2 is that the global scores factorize into local contributions from each node, and, moreover, that in comparing two similar graphs their score difference is identical to quantities which arise when testing conditional independence using cross entropy measures. I shall now show that a similar circumstance arises in a Bayesian approach when globally independent priors are employed. The key feature is that global independence is preserved under updating with complete data (Cowell et al. 1999).

Thus suppose each node $v$ of a graph $g \in \mathcal{G}_n$ has an associated (vector) parameterization $\theta^g_v$ of the conditional probability table of $v$, and a globally independent prior distribution over the parameters $\theta^g := \{\theta^g_v : v \in V\}$. Global independence means that the prior measure factorizes as $d\pi_g(\theta^g) = \prod_v d\pi_g(\theta^g_v)$. Under these conditions the marginal likelihood of the graph $g$ in the light of complete data $D$ is

$$\tilde{L}(g) := p(D|g) = \int p_g(D \mid g, \theta^g) d\pi_g(\theta^g)$$
$$= \prod_v \int_{\theta_v} \prod_{x_v, x_{\mathrm{pa}(v)}} p_g(x_v \mid x_{\mathrm{pa}(v)}, \theta^g_v)^{n(x_v, x_{\mathrm{pa}(v)})} d\pi_g(\theta^g_v).$$
(11)

From (11) we see that the marginal likelihood factorizes into terms, one for each node and it parents. As before, let $g'$ be a graph identical to graph $g$ except for a difference in the parent set of the $X_i$. Then $g'$ will require a different parameterization and associated prior, (see Cowell (1996), Heckerman et al. (1995) for alternative strategies for doing this for Dirichlet priors), but we may take for every node other than $X_i$ the same local parameterization and contribution to the prior as for the graph $g$ (that is, for $X_v \neq X_i$, $\theta^{g'}_v = \theta^g_v$, $P_{g'}(X_v \mid X_{\mathrm{pa}(v:g')}, \theta^{g'}_v) = P_g(X_v \mid X_{\mathrm{pa}(v:g)}, \theta^g_v)$ and $d\pi_g(\theta^g_v) = d\pi_{g'}(\theta^{g'}_v)$). If, furthermore, we take uniform priors over the alternative graphical structures (ie, $P(g) = P(g')$), then after suitable cancellations we obtain the ratio of posterior probabilities given in (12).

$$\frac{p(g'\,|\,D)}{p(g\,|\,D)} = \frac{p(D\,|\,g')}{p(D\,|\,g)} = \frac{\int_{\theta_i^{g'}} \prod_{x_i,x_{\text{pa}(i:g')}} p_{g'}(x_i\,|\,x_{\text{pa}(i:g')},\theta_i^{g'})^{n(x_i,x_{\text{pa}(i:g')})} d\pi_{g'}(\theta_i^{g'})}{\int_{\theta_i^{g}} \prod_{x_i,x_{\text{pa}(i:g)}} p_{g}(x_i\,|\,x_{\text{pa}(i:g)},\theta_i^{g})^{n(x_i,x_{\text{pa}(i:g)})} d\pi_{g}(\theta_i^{g})}. \qquad (12)$$

The decision of a local score driven search to stay with graph $g$ or move to graph $g'$ would depend upon the value of this ratio. Madigan and Raftery (1994) use (the logarithm of) (12) in a Markov chain Monte Carlo based graphical model search procedure, which they apply to model selection and model averaging; see also Madigan and York (1994).

I am not aware of papers applying Bayesian tests of conditional independence to Bayesian network model selection, hence there is not a direct comparison I can make of (12) to results extant in the current literature. (This is not to say there are none; however Bayesian methods — based on comparison of posterior probabilities — for testing for independence in contingency tables do exist, see for example Jeffreys (1961), Good (1965). See also the discussion in Madigan and Raftery (1994).) However, a formal Bayesian approach would consider the following two hypotheses:

$$\begin{aligned}
H_0 \;:\; & p(X_i, X_{\text{pa}(i:g')}\,|\,\theta^{H_0}) d\pi(\theta^{H_0}) \\
\equiv\; & p(X_i\,|\,X_{\text{pa}(i:g')}, \theta_i^{g'}) p(X_{\text{pa}(i:g')}\,|\,\phi_{\text{pa}(i:g')}^{g'}) \\
& d\pi(\theta_i^{g'}) d\pi(\phi_{\text{pa}(i:g')}^{g'}); \\
H_1 \;:\; & p(X_i, X_{\text{pa}(i:g')}\,|\,\theta^{H_1}) d\pi(\theta^{H_1}) \\
\equiv\; & p(X_i\,|\,X_{\text{pa}(i:g)}, \theta_i^{g}) p(X_{\text{pa}(i:g')}\,|\,\phi_{\text{pa}(i:g')}^{g'}) \\
& d\pi(\theta_i^{g}) d\pi(\phi_{\text{pa}(i:g')}^{g'}).
\end{aligned}$$

$H_0$ corresponds to the saturated model, $H_1$ the submodel exhibiting conditional independence, and $\phi$ is a parameterization common to the two hypotheses. Then, from the (possibly equal) priors $P(H_0)$ and $P(H_1)$ and the data $D$, posterior probabilities $P(H_0\,|\,D)$ and $P(H_1\,|\,D)$ are evaluated and compared. It is left to the reader to verify that this leads to (12). Thus if one *were* to do model search based upon local conditional independence tests, then one should use (12) in conjunction with an appropriate decision rule, and then a complete identification of the two approaches — Bayesian score based or Bayesian conditional independence testing — would follow.

## 6 Conclusions

Under the conditions of complete data and given node ordering I have shown that conditional independence tests for searching for Bayesian networks are equivalent to local log-scoring metrics — they are two ways of interpreting the same numerical quantities. It is possible to relax the node-ordering constraint by considering arc reversals in addition to arc removals and additions; then the change in score (which will be local to a pair of nodes) will be a combination of the terms which would be considered using conditional independence tests. However, in the latter case, one would have the extra option of deciding if the conditional independence properties associated with each of the two nodes have been independently locally violated. Thus conditional independence searching *can* be more refined than using scoring metrics when considering arc reversals. However if the individual conditional independence tests were combined into a single test, then the two procedures would again be equivalent under the same decision rules.